\documentclass{article}
\usepackage{ijcai21}

\usepackage{times}

\usepackage{soul}
\usepackage{url}
\usepackage[hidelinks]{hyperref}
\usepackage[utf8]{inputenc}
\usepackage[small]{caption}
\usepackage{graphicx}
\usepackage{amsmath}
\usepackage{booktabs}
\usepackage{microtype}
\usepackage{amsthm}
\usepackage{algorithm}
\usepackage{helvet}
\usepackage{courier}
\usepackage{textcomp}
\usepackage{enumerate}
\usepackage{tabularx}
\usepackage{multirow}
\usepackage{array}
\usepackage{graphics}
\usepackage{graphicx}
\usepackage{subfig}
\usepackage{color}
\usepackage{amsmath}
\usepackage{makecell}
\usepackage{indentfirst}
\usepackage{multirow}
\usepackage{amssymb}
\usepackage{algpseudocode}
\usepackage{comment}
\usepackage{mathrsfs}
\usepackage{extarrows}
\usepackage{booktabs}
\newcommand{\citet}[1]{\citeauthor{#1} \shortcite{#1}}
\newcommand{\citep}{\cite}

\title{TCIC: Theme Concepts Learning Cross Language and Vision for \\Image Captioning}

\author{
Zhihao Fan$^{1}$\and
Zhongyu Wei$^{1,3}$\footnote{Corresponding Author}\and
Siyuan Wang$^{1}$\and
Ruize Wang$^1$\and  \\
Zejun Li$^1$\and
Haijun Shan$^2$\and 
Xuanjing Huang$^1$\and\\
\affiliations
$^1$Fudan University, \\
$^2$Zhejiang Lab,\\
$^3$Research Institute of Intelligent and Complex Systems, Fudan University, China\\
\emails
\{fanzh18,zywei,wangsy18,rzwang18,zejunli20,xjhuang\}@fudan.edu.cn,
workingshan@163.com
}

\begin{document}

\maketitle

\begin{abstract}
Existing research for image captioning usually represents an image using a scene graph with low-level facts (objects and relations) and fails to capture the high-level semantics. In this paper, we propose a \textbf{T}heme \textbf{C}oncepts extended \textbf{I}mage \textbf{C}aptioning (\emph{TCIC}) framework that incorporates theme concepts to represent high-level cross-modality semantics. In practice, we model theme concepts as memory vectors and propose \textbf{T}ransformer with \textbf{T}heme \textbf{N}odes (TTN) to incorporate those vectors for image captioning. Considering that theme concepts can be learned from both images and captions, we propose two settings for their representations learning based on TTN. On the vision side, TTN is configured to take both scene graph based features and theme concepts as input for visual representation learning. On the language side, TTN is configured to take both captions and theme concepts as input for text representation re-construction. Both settings aim to generate target captions with the same transformer-based decoder. During the training, we further align representations of theme concepts learned from images and corresponding captions to enforce the cross-modality learning. Experimental results on MS COCO show the effectiveness of our approach compared to some state-of-the-art models.
\end{abstract}

\section{Introduction}
\noindent Vision and language are two important aspects of human intelligence to understand the world. To bridge vision and language, researchers pay increasing attention to multi-modal tasks. Image captioning~\cite{vinyals2015show}, one of the most widely studied cross-modal topics, aims at constructing a short textual description for the given image. 
\begin{figure}[htbp]
\centering
\includegraphics[width=0.48\textwidth]{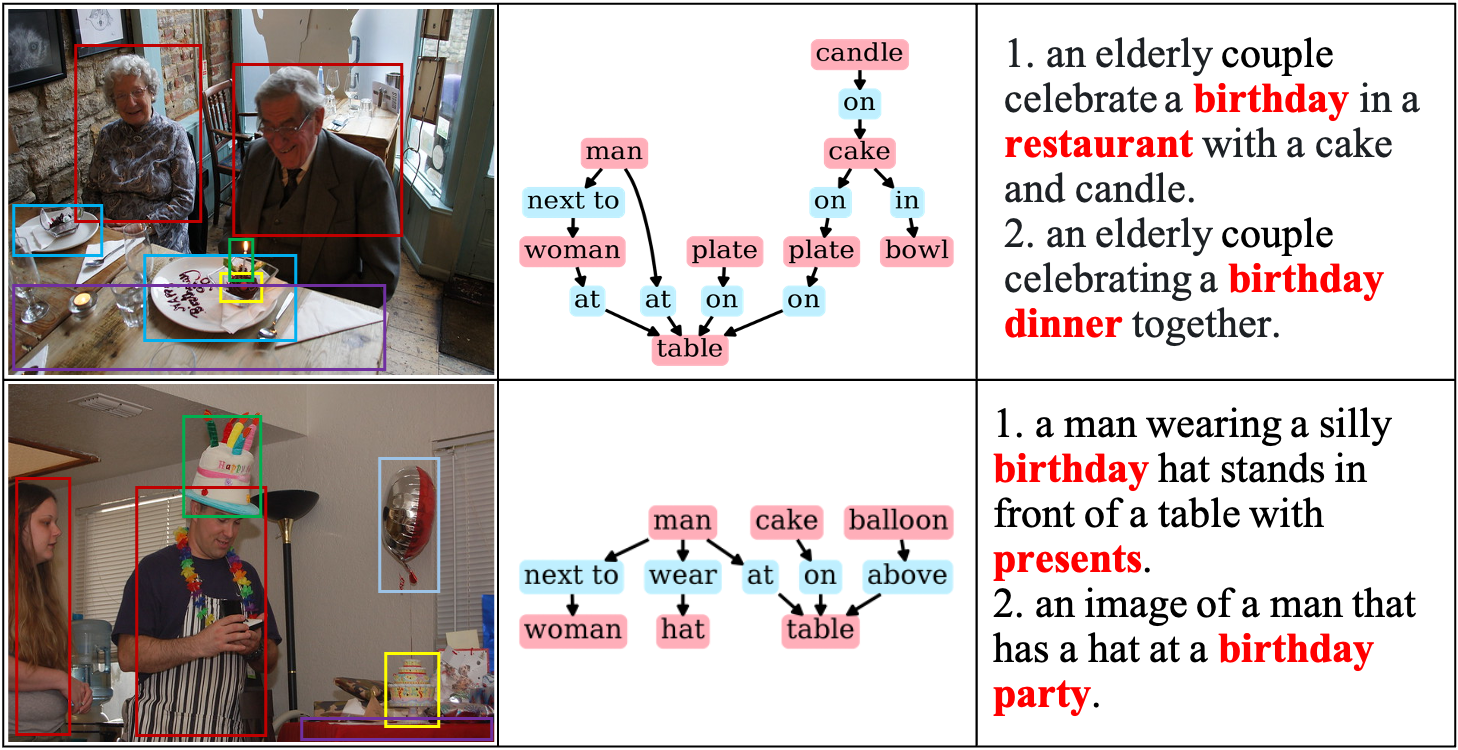}
\centering
\caption{Examples of images, corresponding scene graphs and human-annotated captions. Text in red stands for theme concepts.}
\label{intro_fail_example}
\end{figure}
Existing researches on image captioning usually employ an encoder-decoder architecture~\cite{vinyals2015show,anderson2018bottom} and focus on the problems of image representation learning and cross-modality semantic aligning~\cite{karpathy2015deep,ren2017deep}.

For visual representation learning, the first generation of encoders splits an image into equal-sized regions and extracts CNN-based visual features~\cite{vinyals2015show}. In order to model objects in the image explicitly, Faster-RCNN~\cite{ren2015faster} is proposed to identify bounding boxes of concrete objects. Furthermore, scene graphs~\cite{yao2018exploring,yang2019auto} are introduced to incorporate relations among objects. In a scene graph, region features are extracted as objects and textual features are generated to describe relations. Although positive results have been reported of using scene graphs to represent images for downstream tasks, the semantic gap between visual signals and textual descriptions still exists. 

%, the capability of object detection module is limited by objects annotated in the training corpus and 

%The number of vocabularies in scene graph and description can be seen in Table 1. 

\begin{figure*}%[htbp]
\centering
\includegraphics[width=1.0\textwidth]{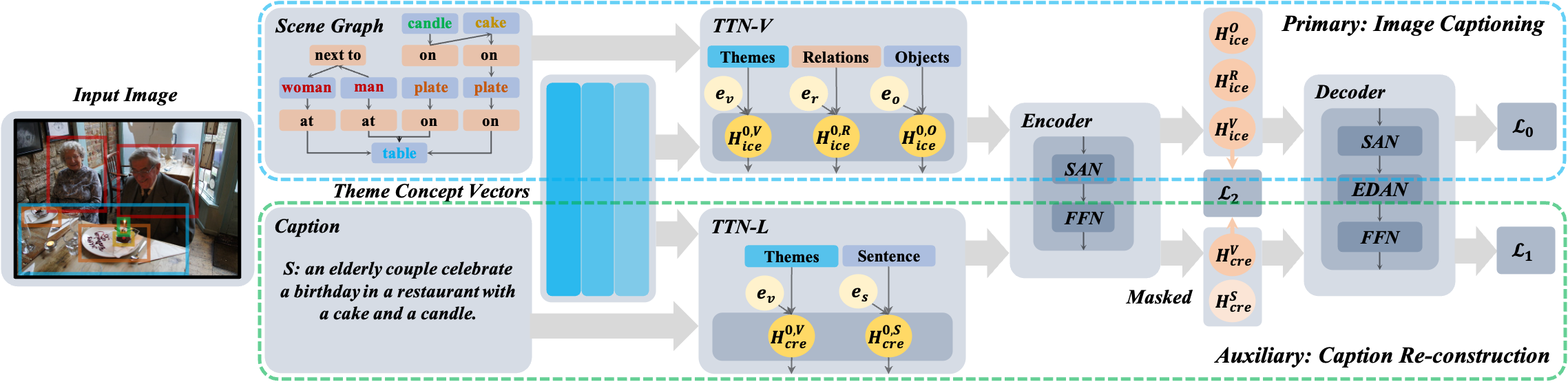}
\centering
\caption{The framework of our proposed model \textbf{T}heme \textbf{C}oncepts extended \textbf{I}mage \textbf{C}aptioner (TCIC). Theme concept vectors are used to represent the high-level cross-modality semantics. And they are updated by interacting with low-level facts in images and tokens in captions inside transformer structure TTN-V and TTN-L via two tasks of image captioning and caption re-construction respectively.}
\label{framework_figure}
\end{figure*}

%It includes training tasks, namely, image captioning(upper) and caption re-construction(down). The inputs of image captioning are objects $\mathcal{O}$, relations $\mathcal{R}$ and theme concepts $\mathcal{V}$, and those of caption re-construction are captions $\mathcal{S}$ and theme concepts $\mathcal{V}$. They share the theme concept vectors and the encoder-decoder parameters, and are also aligned with the representations of theme nodes, namely, $\mathcal{H}_{cre}^{\mathcal{V}}$ and $\mathcal{H}_{ice}^{\mathcal{V}}$.

Figure~\ref{intro_fail_example} presents two examples with images, the corresponding scene graphs and descriptions constructed by humans. On the vision side, scene graph parser identifies some low-level facts of objects (``table'', ``cake'', ``man'', ``candle'', ``hat'', etc.) and relations (``on'', ``next to'', etc.). On the language side, human annotators use some abstract concepts (``birthday'', ``dinner'', ``party'', etc.) of high-level semantics to describe images. This uncovers the semantic gap between the scene graph based visual representation and human language. Therefore, we argue that high-level semantic concepts (also called theme concepts in this paper) can be an extension to scene graphs to represent images. For semantic modeling of images, existing research constructs a list of concept words from descriptions in advance~\cite{you2016image,gan2017semantic,li2019entangled,fan-etal-2019-bridging}, and train an independent module to attach these concept words to images as the guidance for image captioning. Instead of using a pre-defined list of words, we explore to represent theme concepts as shared latent vectors and learn their representations from both images and captions. 
%we use Transformer~\cite{vaswani2017attention} as the backbone structure of our encoder. We Transformer with Theme Node

%A closer look at the example shows that, theme concepts, e.g., ``birthday", are shared in images and can be inferred by different combinations of low-level facts, i.e., \emph{candle on cake}, \emph{cake and balloon on table} and \emph{man wear hat}.

Inspired by the success of Transformer~\cite{NIPS2019_9293,li2019entangled} for image captioning, we propose \textbf{T}ransformer with \textbf{T}heme \textbf{N}ode (TTN) to incorporate theme concepts in the encoder of our architecture. On the vision side, theme concepts can be inferred based on reasoning over low-level facts extracted by scene graphs. For example, ``birthday" can be inferred by different combinations of low-level facts, i.e., \emph{candle on cake}, \emph{cake and balloon on table} and \emph{man wear hat}. Therefore, TTN is configured to integrate three kinds of information (noted as \emph{TTN-V}), namely, objects, relations and theme concepts for visual representation modeling. Inside \emph{TTN-V}, theme concept vectors work as theme nodes and their representations are updated by reasoning over nodes of objects and relations. On the language side, we introduce an auxiliary task named caption re-construction to enable the learning of theme concepts from text corpus. \emph{TTN} is configured to integrate both text information and theme concepts (noted as \emph{TTN-L}). It takes both them concept vectors and captions as input for caption re-construction. Both tasks share the same Transformer-based decoder for caption generation. Besides, we align representations of theme concepts learned from \emph{TTN-V} and \emph{TTN-L} for image-caption pairs to further enforce the cross-modality learning.

% To distinguish different kinds of nodes, we utilize group embedding to learn group-wise representations. % Besides self-attention, group attention mechanism is introduced to better incorporate information from different groups of nodes for better representation learning. 

We conduct experiments on MS COCO~\cite{lin2014microsoft}. Both offline and online testings show the effectiveness of our model compared to some state-of-the-art approaches in terms of automatic evaluation metrics. We further interpret the semantics of theme concepts via their related objects in the image and words in the caption. Results show that theme concepts are able to bridge the semantics of language and vision to some extent. 

\section{Related Work}

\label{SectionRelatedWork}
%Research on the cross modality of vision and language has attracted increasing attention recently. 
% Early works in the cross modality of vision and language tend to use rules or templates for generation. 
\noindent Motivated by the encoder-decoder architecture, models produce texts from from image have many variants and improvement~\cite{you2016image,anderson2018bottom,yang2019auto,fan2018question,wang2020storytelling,fan-etal-2019-bridging}. In image captioning,~\citet{fang2015captions},~\citet{you2016image},~\citet{Gan_2017_CVPR} and~\citet{liu2019aligning} pre-define a list of words as semantic concepts for image captioning. ~\citet{you2016image} employ an attention mechanism over word-level concepts to enhance the generator. ~\citet{li2019entangled} propose
to simultaneously exploit word concepts and visual information in decoder. 
Some researchers also explore Transformer-based model for image captioning~\cite{li2019entangled,liu2019aligning,huang2019attention,NIPS2019_9293,cornia2020m2}. ~\citet{NIPS2019_9293} propose to better model the spatial relations between detected objects through geometric attention. ~\citet{huang2019attention} extend self-attention to determine the relevance between attention outputs and query objects for refinement. ~\citet{cornia2020m2} introduce persist memory to self-attention key-value pairs as prior knowledge to enhance the generation.

The most relevant works to our research are \emph{HIP}~\cite{yao2019hierarchy}, \emph{MMT}~\cite{cornia2020m2} and \emph{SGAE}~\cite{yang2019auto}. Both \emph{HIP}~\cite{yao2019hierarchy} and our work explore to model structure information of images. \emph{HIP} utilizes Mask R-CNN to identify instances in region via image segmentation. Our model is built on top of scene graph and is able to identify high-level semantics of the whole image. Our assumption is that objects spread in different corners of the image can also express the high-level semantics together, therefore, our model utilizes the low-level facts in the scene graph to explore high-level semantics. All of \emph{SGAE}, \emph{MMT} and our work utilize memory vectors, but the memory vectors of \emph{SGAE} and \emph{MMT} are fixed and non-interactive, which means that their memory vectors only provide prior information and are unable to actively learn the cross-modality theme concepts. In our framework theme concept vectors are learned by interacting with low-level facts in images and tokens in captions based on Transformer.

\section{\textbf{T}heme \textbf{C}oncepts Extended \textbf{I}mage \textbf{C}aptioning}

\noindent  The overall framework of our model \textbf{T}heme \textbf{C}oncepts extended \textbf{I}mage \textbf{C}aptioning (\emph{TCIC}) is shown in Figure~\ref{framework_figure}. We model theme concepts as shared memory vectors ($\mathcal{V}$) and learn their representations inside \textbf{T}ransformer with \textbf{T}heme \textbf{N}ode (TTN) from both images and captions via two tasks. The primary task is image captioning (upper). TTN is configured as TTN-V to learn the visual representation for image captioning. It takes scene graph features (objects $\mathcal{O}$ and relations $\mathcal{R}$) and theme concepts $\mathcal{V}$ as input.

\begin{equation}
    \begin{small}
        \begin{aligned}
\mathcal{G}_{\theta}\big(\mathcal{V}, \mathcal{O}, \mathcal{R}\big)\rightarrow \mathcal{S} 
        \label{our_ic_task}
        \end{aligned}
    \end{small}
\end{equation}

The auxiliary task is caption re-construction (bottom). \emph{TTN} is configured as \emph{TTN-L} to learn the text representation for caption re-construction. It takes both textual features $\mathcal{S}$ and theme concepts $\mathcal{V}$ as input. 
\begin{equation}
    \begin{small}
        \begin{aligned}
\mathcal{G}_{\theta}\big(\mathcal{V}, \mathcal{S}\big)\rightarrow \mathcal{S} 
        \label{our_sr_task}
        \end{aligned}
    \end{small}
\end{equation}

%The primary task image captioning is to build $\mathcal{G}_\theta$ to generate the target caption $\mathcal{S}$. To better infer the high-level semantics of image, instead of only extracting the objects $\mathcal{O}$ and relations $\mathcal{R}$ in the image scene graph $\mathcal{SG}$ as the input features, we also add theme nodes $\mathcal{V}$ to capture the concept knowledge from $\mathcal{SG}$ as Eq.~(\ref{our_ic_task}), where $(\mathcal{V},\mathcal{O},\mathcal{R})$ constitute a new super graph on top of $\mathcal{SG}$.

%We observe that the semantic concepts are presented in text corpus, but caption in $\mathcal{S}$ is not directly accessible to theme concept vectors $\mathcal{V}$ in the primary task image captioning. To improve the capability of $\mathcal{V}$ in concept representation, we propose the auxiliary task caption re-construction as Eq.~(\ref{our_sr_task}) where $\mathcal{V}$ directly interact with $\mathcal{S}$ to learn the knowledge of concepts. 

%It includes two training tasks, namely, image captioning (upper) and caption re-construction (down). The inputs of image captioning are objects $\mathcal{O}$, relations $\mathcal{R}$ and theme concepts $\mathcal{V}$, and those of caption re-construction are captions $\mathcal{S}$ and theme concepts $\mathcal{V}$. 

Note that both TTN-V and TTN-L share the same architecture and parameters. Besides, we use the same decoder for caption generation in both tasks. During the training, representations of theme nodes learned in TTN-V ($\mathcal{H}_{ice}^{\mathcal{V}}$) and TTN-L ($\mathcal{H}_{cre}^{\mathcal{V}}$) are further aligned.

\subsection{Image Encoder with TTN-V}
\noindent  The image encoder utilizes TTN-V to incorporate both scene graph features and theme concept vectors for image representation learning. 

\paragraph{Inputs of TTN-V.}
We extract objects $\mathcal{O}$ and relations $\mathcal{R}$ as the scene graph $\mathcal{SG}$ of the image. Then, theme concept vectors are used to extend $\mathcal{SG}$. Moreover, we employ multiple theme concept vectors to model image theme concepts from different aspects. Therefore, our image captioning inputs $\mathcal{I}_{ic}$ have three groups:
\begin{equation}
    \begin{small}
        \begin{aligned}
            \mathcal{I}_{\emph{ic}}=\big[\mathcal{V}, \mathcal{O}, \mathcal{R}\big]
        \end{aligned}
    \end{small}
\end{equation}

The three groups of nodes play different roles for visual semantics modeling, we utilize group embeddings, namely, $\big\{e_{o},e_{r},e_{v}\big\}\in \mathbb{R}^{d}$, to distinguish them as Eq.~(\ref{embedding}).
\begin{equation}
    \begin{small}
        \begin{aligned}
            \mathcal{H}^{\emph{0}}_{\emph{ice}}&=\left\{
            \begin{array}{ll}·
                \textbf{Emb}(v_{i})+e_v, &v_{i}\in \mathcal{V}\vspace{1ex} \\
                \textbf{W}_{\textbf{o}}\big[f_{i},p_{i}\big]+e_{o}, &o_{i}\in \mathcal{O}\vspace{1ex}  \\
                \textbf{Emb}(r_{i})+e_r, &r_{i}\in \mathcal{R}
            \end{array}
            \right.\label{embedding}  
        \end{aligned}
    \end{small}
\end{equation}
where $\textbf{W}_{\text{o}}\in \mathbb{R}^{d\times (d_{o}+5)}$ is a trainable matrix, $d_{o}$ is the region feature dimension and $d$ is the hidden dimension of our encoder. $f_{i}\in\mathbb{R}^{d_{o}}$ is the region context feature of object $o_{i}$. $p_{i}=\big(\frac{x_{1}}{w},\frac{y_{1}}{h}, \frac{x_{2}}{w},\frac{y_{2}}{h},\frac{(y_{2}-y_{1})(x_{2}-x_{1})}{wh}\big)$, where $(x_{1},y_{1})$ and $(x_{2},y_{2})$ denote the coordinate of the bottom-left and top-right corner of object $o_{i}$, while $w$ and $h$ are the width and height of the input image. $\textbf{Emb}$ is the embedding function of theme nodes and relations.

\begin{table*}
\begin{center}
\resizebox{1.0\textwidth}{!}{
%\textcolor{blue}{}%
\begin{tabular}{l|cccccc|cccccc}
\midrule[1.0pt]
\multirow{2}{*}{Model} &\multicolumn{6}{c|}{Cross Entropy}  &\multicolumn{6}{c}{RL} \\
&B-1 &B-4 &M &R &C &S &B-1 &B-4 &M &R &C &S \\
\midrule[1.0pt]
&\multicolumn{12}{c}{Single Model} \\
\midrule[1.0pt]
% \emph{Hard-Attention}~\cite{xu2015show} &25.0 &- &23.0 &- &19.9 &- &18.5 &-\\
%\emph{SCA-CNN}~\cite{chen2017sca} &31.1 &- &25.0 &- &22.3 &- &19.5 &-\\
\emph{NIC}~\cite{vinyals2015show} &- &29.6 &25.2 &52.6 &94.0 &- &- &31.9 &25.5 &54.3 &106.3 &- \\
\emph{Up-Down}~\cite{anderson2018bottom} &77.2 &36.2 &27.0 &56.4 &113.5 &20.3 &79.8 &36.3 &27.7 &56.9 &120.1 &21.4 \\
\emph{GCN-LSTM}~\cite{yao2018exploring} &77.3 &36.8 &27.9 &57.0 &116.3 &20.9 &80.5 &38.2 &28.5 &58.3 &127.6 &22.0 \\
%Standard Transformer 113.21 21.04 75.60 34.58 27.79 56.02
\emph{TOiW}~\cite{NIPS2019_9293} &76.6 &35.5 &27.8 &56.6 &115.4 &21.2 &80.5 &38.6 &28.7 &58.4 &128.3 &22.6 \\
% \emph{ETA}~\cite{li2019entangled} &77.3 &- &- &37.1 &28.2 &57.1 &117.9 &21.4 &81.5 &39.3 &28.8 &58.9 &126.6 &22.7 \\
% \emph{AoA}~\cite{huang2019attention} &77.4 &37.2  &28.4 &57.5 &119.8 &21.3  &80.2 &38.9 &29.2 &\textbf{58.8} &129.8 &22.4 \\
\emph{SGAE}~\cite{yang2019auto} &77.6 &36.9 &27.7 &57.2 &116.7 &20.9 &80.8 &38.4 &28.4 &58.6 &127.8 &22.1 \\
\emph{MMT}~\cite{cornia2020m2} &- &- &- &- &- &- &80.8 &39.1 &29.2 &58.6 &131.2 &\textbf{22.6} \\
% \emph{CAAG}&- &- &- &- &- &- &- &- &- &39.4 &29.5 &59.2 &132.2 &22.8 \\
% \emph{X-LAN} &78.0 &62.3 &48.9 &38.2 &\textbf{28.8} &58.0 &\textbf{122.0} &\textbf{21.9} &80.8 &65.6 &51.4 &39.5 &29.5 &59.2 &132.0 &23.4 \\
% \emph{X-Trans}~\cite{} &77.3 &61.5 &47.8 &37.0 &\textbf{28.7} &57.5 &120.0 &\textbf{21.8} &80.9 &39.7 &29.5 &59.1 &132.8 &23.4 \\
\midrule[0.5pt]
\emph{TCIC}\ [Ours] &\textbf{78.1} &\textbf{38.3} &\textbf{28.5} &\textbf{58.0} &\textbf{121.0} &\textbf{21.6} &\textbf{80.9} &\textbf{39.7} &\textbf{29.2} &\textbf{58.6} &\textbf{132.9} &22.4 \\
\midrule[1.0pt]
&\multicolumn{12}{c}{Ensemble Model} \\
\midrule[1.0pt]
\emph{SGAE$^{\Sigma}$}~\cite{yang2019auto} &- &- &- &- &- &- &81.0 &39.0 &28.4 &58.9 &129.1 &22.2\\
\emph{GCN-LSTM$^{\Sigma}$}~\cite{yao2018exploring} &77.4 &37.1 &28.1 &57.2 &117.1 &21.1 &80.9 &38.3 &28.6 &58.5 &128.7 &22.1 \\
% \emph{ETA*}~\cite{li2019entangled} &77.6 &- &- &37.8 &28.4 &57.4 &119.3 &21.6 &81.5 &39.9 &28.9 &\textbf{59.0} &127.6 &22.6 \\
%\emph{AoANet$^{\Sigma}$}~\cite{huang2019attention} &78.7 &38.1 &28.5 &58.2 &122.7 &21.7 &81.6 &40.2 &29.3 &59.4 &132.0 &22.8 \\
\emph{HIP$^{\Sigma}$}~\cite{yao2019hierarchy} &- &38.0 &28.6 &57.8 &120.3 &21.4  &- &39.1 &28.9 &59.2 &130.6 &22.3 \\
\emph{MMT$^{\Sigma}$}~\cite{cornia2020m2} &- &- &- &- &- &- &81.6 &39.8 &29.5 &59.2 &133.2 &\textbf{23.1} \\
% \emph{X-LAN*} &78.8 &63.4 &49.9 &39.1 &29.1 &58.5 &\textbf{124.5} &22.2 &81.6 &66.6 &52.3 &40.3 &29.8 &59.6 &133.7 &23.6 \\
% \emph{X-Transformer*} &77.8 &62.1 &48.6 &37.7 &29.0 &58.0 &122.1 &21.9 &81.7 &40.7 &29.9 &59.7 &135.3 &23.8 \\
\midrule[0.5pt]
\emph{TCIC$^{\Sigma}$}\ [Ours] &\textbf{78.8} &\textbf{39.1} &\textbf{29.1} &\textbf{58.5} &\textbf{123.9} &\textbf{22.2} &\textbf{81.8} &\textbf{40.8} &\textbf{29.5} &\textbf{59.2} &\textbf{135.3} &22.5 \\
\midrule[1.0pt]
\end{tabular}
}
\end{center}
\caption{Overall performance of MS COCO offline testing. B-1, B-4, R, M, C and S are short for BLEU-1, BLEU-4, ROUGE, METEOR, CIDEr-D and SPICE, respectively. ${\Sigma}$ means ensemble model. Numbers in \textbf{bold} denote the best performance in each column.} %The difference of all metrics between \emph{NIC+WC+WA+RL} and those models whose name begin with \emph{NIC} except \emph{NIC+WC(GT)} is statistically significant($p<0.05$)}.
\label{OverallPerformance}
\end{table*}

% \paragraph{Encoder-Decoder}
% We use Transformer~\cite{vaswani2017attention} as our backbone.  There are three types of sublayer in Transformer, namely, Self-Attention Network (\textbf{SAN}), Feed-Forward Network (\textbf{FFN}) and Encoder-Decoder Attention Network (\textbf{EDAN}). % Transformer encoder consists of SAN and FFN, and decoder includes SAN, EDAN and FFN. 

\paragraph{Structure of Image Encoder.} 
Each encoder layer in the image encoder includes Self-Attention Network $\textbf{SAN}$ and Feed-Forward Network $\textbf{FFN}$. It takes $\mathcal{H}^{\mathit{l-1}}_{\mathit{ice}}$ as inputs.

The key of \textbf{SAN} is multihead attention $\textbf{MHA}$ as Eq.~(\ref{mha-layer}). % and we would inject the structure knowledge of $\mathcal{SG}$ into $\textbf{MHA}$. We first describe  $\textbf{MHA}$.
\begin{equation}
    \begin{small}
        \begin{gathered}
            \textbf{MHA}(\text{Q},\text{K},\text{V})= \textbf{W}_\textbf{o} \big[\text{head}_{\text{1}},\cdots,\text{head}_{\text{m}}\big] \vspace{1ex} ~\label{mha-layer} \\
            \text{head}_{\text{i}}=\textbf{Attn}\big(\textbf{W}_{\textbf{q}}\text{Q}, \textbf{W}_{\textbf{k}}\text{K},\textbf{W}_{\textbf{v}}\text{V}\big)
        \end{gathered})
    \end{small}
\end{equation}
where $\text{m}$ is the number of attention heads, and $\textbf{W}_{\textbf{o}}\in\mathbb{R}^{d\times d}$ is the trainable output matrix.
Moreover, the attention function $\textbf{Attn}$ maps a query and a set of key-value pairs to an output:
\begin{equation}
    \begin{small}
        \begin{gathered}
            \textbf{Attn}(\text{Q}, \text{K}, \text{V}) = \textbf{Softmax}\bigg(\frac{\text{Q}^{\text{T}}\text{K}}{\sqrt{d_{k}}}\bigg)\text{V}~\label{self-attention}
        \end{gathered}
    \end{small}
\end{equation}
where the queries $\text{Q}\in\mathbb{R}^{d_{k}\times n_{q}}$, keys $\text{K}\in\mathbb{R}^{d_{k}\times n_{k}}$ and values $\text{V}\in\mathbb{R}^{d_{k}\times n_{k}}$, $d_{k}$ is the attention hidden size, $n_{q}$ and $n_{k}$ are the number of query and key, respectively.

%\paragraph{Self-Attention}
$\mathcal{SG}$ in the encoder input has the inherent structure, i.e., $(o_{i}, r_{j}, o_{k})$. Thus we adopt hard mask for the triplets in $\mathcal{SG}$ to inject the structure knowledge into $\textbf{MHA}$. %as part of our model input, is not plain text as in the original Transformer~\cite{vaswani2017attention} but has structure. 
In detail, a matrix $\text{M}\in\mathbb{R}^{|\mathcal{I}_{\mathit{ic}}|\times |\mathcal{I}_{\mathit{ic}}|}$ is initialized with all $0$. For any object $o_{i}\in\mathcal{O}$ and any relation $r_{j}\in \mathcal{R}$, if there does not exist any object $o_{k}\in \mathcal{O}$, such that $(o_{i},r_{j},o_{k})\in\mathcal{SG}$, then we set $\text{M}_{i,j}=-\infty$. Following Eq.~(\ref{mha-layer}), we add $\text{M}$ to $\textbf{Attn}$ and get $\textbf{MAttn}$.
% The matrix $\text{M}$ gets involved in attention computation, through which the attention between objects and relations are limited by triplets in $\mathcal{SG}$. 
\begin{equation}
    \begin{small}
        \begin{gathered}
            \textbf{MAttn}(\text{Q},\text{K},\text{V},\text{M})=\textbf{Softmax}\bigg(\text{M}+\frac{\text{Q}^{\text{T}}\text{K}}{\sqrt{d_{k}}}\bigg)\text{V}~\label{mmha-layer}
        \end{gathered}
    \end{small}
    % \label{Mask-attention-layer}
\end{equation}

Through replacing $\textbf{Attn}$ with $\textbf{MAttn}$, we build the masked multihead attention $\textbf{MMHA}$.% in Eq.~(\ref{attention-layer}).
The details of our image captioning encoder layer is shown in Eq.~(\ref{ic-encoder-layer})
\begin{equation}
    \begin{small}
        \begin{aligned}
            \widehat{\mathcal{H}}^{\mathit{l}}_{\mathit{ice}}&=\textbf{LN}\big(\mathcal{H}^{\mathit{l-1}}_{\mathit{ice}}+\textbf{MMHA}\big(\mathcal{H}^{\mathit{l-1}}_{\mathit{ice}},\mathcal{H}^{\mathit{l-1}}_{\mathit{ice}},\mathcal{H}^{\mathit{l-1}}_{\mathit{ice}}\big)\big) \\  
            \mathcal{H}^{\mathit{l}}_{\mathit{ice}}&=\textbf{LN}\big(\widehat{\mathcal{H}}^{\mathit{l}}_{\mathit{ice}}+\textbf{W}_{2}\ \textbf{ReLU}\big(\textbf{W}_{1}\widehat{\mathcal{H}}^{\mathit{l}}_{\mathit{ice}}\big)\big)
            \label{ic-encoder-layer}
        \end{aligned}
    \end{small}
\end{equation}
where $\textbf{LN}$ is LayerNorm. % $\textbf{W}_{1}$ and $\textbf{W}_{2}$ are trainable parameters.

Through image captioning encoding, we get the outputs $\mathcal{H}^{\mathcal{E}}_{\mathit{ice}}$. It consists of $\mathcal{H}^{\mathcal{V}}_{\mathit{ice}}$, $\mathcal{H}^{\mathcal{O}}_{\mathit{ice}}$ and $\mathcal{H}^{\mathcal{R}}_{\mathit{ice}}$, corresponding to $\mathcal{V}$, $\mathcal{O}$ and $\mathcal{R}$.

\subsection{Caption Encoder with TTN-L}
\noindent The caption encoder utilizes TTN-L to incorporate both textural features and theme concept vectors for caption re-construction. 
\paragraph{Inputs of TTN-L.}
We concatenate the target sentence $\mathcal{S}$ and theme nodes $\mathcal{V}$ as inputs of the caption encoder $\mathcal{I}_{\emph{sr}}$ in Eq.~(\ref{sr_input}). 
\begin{equation}
    \begin{small}
        \begin{gathered}
            \mathcal{I}_{\emph{sr}}=\big[\mathcal{V},\mathcal{S}]~\label{sr_input}
        \end{gathered}
    \end{small}
\end{equation}

We also use group embeddings, $\big\{e_{v},e_s\big\}$, to distinguish theme nodes and words of captions in the embedding function as Eq.~(\ref{sr_embedding}).
% The embedding function for $\mathcal{I}_{\emph{sr}}$ follows the Equation~\ref{sr_embedding}, where we utilize group embeddings, namely, $\big\{e_{v},e_s\big\}\in \mathbb{R}^{d}$, to distinguish theme nodes and sentences. 
\begin{equation}
    \begin{small}
        \begin{aligned}
            \mathcal{H}^{\emph{0}}_{\mathit{cre}}&=\left\{
            \begin{array}{ll}
                \mathbf{Emb}(v_{i})+e_v, &v_{i}\in \mathcal{V}\vspace{1ex} \\
                \mathbf{Emb}(s_{i})+\mathbf{Emb}_{\mathit{p}}(s_{i})+e_s, &s_{i}\in \mathcal{S}
            \end{array}
            \right.\label{sr_embedding}
        \end{aligned}
    \end{small}
\end{equation}
where $\textbf{Emb}$ is the embedding function of theme nodes and words, and $\textbf{Emb}_{\mathit{p}}$ is the position embedding following~\citet{vaswani2017attention}. 

\paragraph{Structure of Caption Encoder.}
The caption encoder is the same as the image encoder except that it uses \textbf{MHA} instead of \textbf{MMHA}, but they share the same parameters. Taking $\mathcal{H}^{\mathit{0}}_{\mathit{cre}}$ as the input of the caption encoder, we get the outputs $\mathcal{H}^{\mathcal{E}}_{\mathit{cre}}$. It consists of $\mathcal{H}^{\mathcal{V}}_{\mathit{cre}}$ and $\mathcal{H}^{\mathcal{S}}_{\mathit{cre}}$, corresponding to $\mathcal{V}$ and $\mathcal{S}$.
% Each encoder layer of caption re-construction encoder includes $\textbf{SAN}$ and $\textbf{FFN}$, and takes $\mathcal{H}^{\mathit{l-1}}_{\mathit{cre}}$ as inputs.
% In the $l$-th caption re-construction encoder layer, the inputs $\mathcal{H}^{\mathit{l-1}}_{\mathit{sr}}$ go through $\textbf{SAN}$ and $\textbf{FFN}$. 
% \begin{equation}
%     \begin{small}
%         \begin{aligned}
%              \widehat{\mathcal{H}}^{\mathit{l}}_{\mathit{cre}}&=\textbf{LN}\big(\mathcal{H}^{\mathit{l-1}}_{\mathit{cre}}+\textbf{MHA}\big(\mathcal{H}^{\mathit{l-1}}_{\mathit{cre}},\mathcal{H}^{\mathit{l-1}}_{\mathit{cre}},\mathcal{H}^{\mathit{l-1}}_{\mathit{cre}}\big)\big) \\  
%              \mathcal{H}^{\mathit{l}}_{\mathit{cre}}&=\textbf{LN}\big(\widehat{\mathcal{H}}^{\mathit{l}}_{\mathit{cre}}+\textbf{W}_{2}\ \textbf{ReLU}\big(\textbf{W}_{1}\widehat{\mathcal{H}}^{\mathit{l}}_{\mathit{cre}}\big)\big)~\label{ffn}
%         \end{aligned}
%     \end{small}
% \end{equation}
% After the encoding in caption re-construction, we get the outputs $\mathcal{H}^{\mathcal{E}}_{\mathit{cre}}$. It consists of $\mathcal{H}^{\mathcal{V}}_{\mathit{cre}}$ and $\mathcal{H}^{\mathcal{S}}_{\mathit{cre}}$, corresponding to the $\mathcal{V}$ and $\mathcal{S}$.

\subsection{Decoder for Caption Generation}
% \paragraph{Decoder with $\mathcal{H}^{\mathcal{E}}_{\mathit{ice}}$}  
\noindent We use the same decoder for both image captioning and caption re-construction. The embedding of decoder is initialized with $\mathcal{H}^{\mathit{0}}_{\mathit{icd}}$, which contains word embedding and position embedding following~\cite{vaswani2017attention}. 

In the $l$-th decoder layer, the inputs $\mathcal{H}^{\mathit{l-1}}_{\mathit{d}}$ go through $\textbf{SAN}$, $\textbf{EDAN}$ and $\textbf{FFN}$. 
\begin{equation}
    \begin{small}
        \begin{aligned}
            \bar{\mathcal{H}}^{\mathit{l}}_{\mathit{d}}&=\textbf{LN}\big(\mathcal{H}^{\mathit{l-1}}_{\mathit{d}}+\textbf{MHA}\big(\mathcal{H}^{\mathit{l-1}}_{\mathit{d}},\mathcal{H}^{\mathit{l-1}}_{\mathit{d}},\mathcal{H}^{\mathit{l-1}}_{\mathit{d}}\big)\big) \\  
            \widehat{\mathcal{H}}^{\mathit{l}}_{\mathit{d}}&=\textbf{LN}\big(\bar{\mathcal{H}}^{\mathit{l}}_{\mathit{d}}+\textbf{MHA}\big(\bar{\mathcal{H}}^{\mathit{l}}_{\mathit{d}},\mathcal{H}^{\mathcal{E}}_{\mathit{e}},\mathcal{H}^{\mathcal{E}}_{\mathit{e}}\big)\big) \\  
            \mathcal{H}^{\mathit{l}}_{\mathit{d}}&=\textbf{LN}\big(\widehat{\mathcal{H}}^{\mathit{l}}_{\mathit{d}}+\textbf{W}_{2}\ \textbf{ReLU}\big(\textbf{W}_{1}\widehat{\mathcal{H}}^{\mathit{l}}_{\mathit{d}}\big)\big)~\label{ffn}
        \end{aligned}
    \end{small}
\end{equation}
It is worth noting that, there is a difference between image captioning and caption re-construction in $\textbf{EDAN}$. For image captioning, $\bar{\mathcal{H}}^{\mathit{l}}_{\mathit{d}}$ are able to attend to all key-value pairs $\mathcal{H}^{\mathcal{E}}_{\mathit{ice}}$, but in caption re-construction, only the outputs of theme concept vectors, $\mathcal{H}^{\mathcal{V}}_{\mathit{cre}}$, are visible. Through this method, the theme concept vectors are encouraged to better capture the concept knowledge in captions $\mathcal{S}$. 
\begin{equation}
    \begin{small}
        \begin{aligned}
            \mathcal{H}^{\mathcal{E}}_{\mathit{e}}&=\left\{
            \begin{array}{ll}
                \mathcal{H}^{\mathcal{E}}_{\mathit{ice}},\quad \text{image captioning}\vspace{1ex} \\
                \mathcal{H}^{\mathcal{V}}_{\mathit{cre}},\quad \text{caption re-construction}
            \end{array}
            \right.\label{sr_embedding1}
        \end{aligned}
    \end{small}
\end{equation}

We get the outputs $\mathcal{H}^{\mathcal{D}}_{\mathit{d}}$ after decoding. At last, $\mathcal{H}^{\mathcal{D}}_{\mathit{d}}$ is utilized to estimate the word distribution as Eq.~(\ref{word_prediction}).
\begin{equation}
    \begin{small}
        \begin{aligned}
            \textbf{P}(\mathcal{S})=\text{Softmax}\big(\textbf{W}_{d}\mathcal{H}^{\mathcal{D}}_{\mathit{d}}+b_{d}\big)~\label{word_prediction}
        \end{aligned}
    \end{small}
\end{equation}

\subsection{Overall Training}
\noindent Our training has two phases, cross-entropy based training and RL based training. For cross-entropy based training, the objective is to minimize the negative log-likelihood of $\mathcal{S}$ given $\mathcal{I}_{\mathit{ic}}$, $\mathcal{L}_{0}$, and the negative log-likelihood of $\mathcal{S}$ given $\mathcal{I}_{\mathit{sr}}$, $\mathcal{L}_{1}$. To align the learning of theme concept vectors cross language and images, we add $\mathcal{L}_{2}$ to minimize the distance between $\mathcal{H}^{\mathcal{V}}_{\mathit{ice}}$ and $\mathcal{H}^{\mathcal{V}}_{\mathit{cre}}$.
\begin{equation}
    \begin{small}
        \begin{gathered}
            \mathcal{L}_{0}=-\log \textbf{P}\big(\mathcal{S}|\mathcal{I}_{\mathit{ic}},\mathcal{G}_{\theta}\big),\ \mathcal{L}_{1}=\log \textbf{P}\big(\mathcal{S}|\mathcal{I}_{\mathit{sr}},\mathcal{G}_{\theta}\big) \\
             \mathcal{L}_{2}=\big\|\mathcal{H}^{\mathcal{V}}_{\mathit{ice}}/\|\mathcal{H}^{\mathcal{V}}_{\mathit{ice}}\|_{2}^{2}-\mathcal{H}^{\mathcal{V}}_{\mathit{cre}}/\|\mathcal{H}^{\mathcal{V}}_{\mathit{cre}}\|_{2}^{2}\big\|_{2}^{2}  \\
            \mathcal{L}=\mathcal{L}_{0}+\lambda_{1}\mathcal{L}_{1}+\lambda_{2}\mathcal{L}_{2}
        \end{gathered}
    \end{small}
\end{equation}
where $\lambda_{1}$ and $\lambda_{2}$ are the factors to balance image captioning, caption re-construction and theme node alignments.

The next phase is to use reinforcement learning to finetune $\mathcal{G}_{\theta}$. Following~\citet{rennie2016self}, we use the CIDEr score as the reward function $r$ because it well correlates with the human judgment in image captioning~\cite{vedantam2015cider}. Our training target is to maximize the expected reward of the generated sentence $\widehat{\mathcal{S}}$ as Eq.~(\ref{rl-target}).
\begin{equation}
    \begin{small}
        \begin{aligned}
            \max_{\theta}L_{\theta}= \max_{\theta}\mathbb{E}_{\widehat{S}\sim \mathcal{G}_{\theta}}\big[r(\widehat{\mathcal{S}})\big]\label{rl-target}
        \end{aligned}
    \end{small}
\end{equation}

Then following the reinforce algorithm, we generate $K$ sentences, $\widehat{\mathcal{S}}_{\mathit{1}},\cdots,\widehat{\mathcal{S}}_{\mathit{K}}$, with the random sampling decoding strategy and use the mean of rewards as the baseline. The final gradient for one sample is thus in Eq.~(\ref{rl}).
\begin{equation}
    \begin{small}
        \begin{gathered}
       \bigtriangledown_\theta L_{\theta}=-\frac{1}{K}\sum_{k}\big(r(\widehat{S}_{k})-b_k\big)\bigtriangledown_\theta \log p(\widehat{\mathcal{S}}_{k}|\mathcal{I}_{\mathit{ic}},\mathcal{G}_{\theta}) \\
      b_k=\frac{1}{K-1}\sum_{j\ne k}r\big(\widehat{\mathcal{S}}_{j}\big) \label{rl}
        \end{gathered}
    \end{small}
\end{equation}

During prediction, we decode with beam search, and keep the sequence with highest predicted probability among those in the last beam.

\section{Experiment and Results}

\begin{table}
\begin{center}
\resizebox{0.46\textwidth}{!}{
%\textcolor{blue}{}%
\begin{tabular}{c|ccccccc}
\midrule[1.0pt]
% \multirow{2}{*}{Model} 
Model& &B-1 &B-4 &M &R &C\\
\midrule[1.0pt]
% \midrule[1.0pt]
% \multirow{2}{*}{\emph{GCN-LSTM$^{\Sigma}$}} &c5 &- &38.7 &28.5 &  &125.3 & \\
% &c40 &- &69.7 &37.6 & &126.5 & \\
% \midrule[1.0pt]
\multirow{2}{*}{\emph{SGAE$^{\Sigma}$}} &c5 &81.0 &38.5 &28.2 &58.6 &123.8  & \\
&c40 &95.3 &69.7 &37.2 &73.6 &126.5 \\
\midrule[1.0pt]
\multirow{2}{*}{\emph{HIP$^{\Sigma}$}} &c5 &81.6 &39.3 &28.8 &59.0 &127.9  & \\
&c40 &95.9 &71.0 &38.1 &74.1 &130.2 \\
%\midrule[1.0pt]
%\multirow{2}{*}{\emph{MMT$^{\Sigma}$}} &c5 &81.6 &39.7 &29.4 &59.2 &129.3  \\
%&c40 &96.0 &72.8 &39.0 &74.8 &132.1 \\
\midrule[1.0pt]
% \multirow{2}{*}{\emph{X-LAN$^{\Sigma}$}} &c5 &81.3 &40.3 &29.2 & &129.3  & \\
% &c40 &95.3 &69.7 &37.2 & &131.4 \\
%\midrule[1.0pt]
%\multirow{2}{*}{\emph{ETA*}} &c5 &81.2 &38.9 &28.6 &122.1 \\
%&c40 &95.0 &69.7 &38.0 &124.4 \\
% \midrule[1.0pt]
\multirow{2}{*}{\emph{TCIC$^{\Sigma}$}} &c5 &\textbf{81.8} &\textbf{40.0} &\textbf{29.2} &\textbf{59.0} &\textbf{129.5} & \\
&c40 &\textbf{96.0} &\textbf{72.9} &\textbf{38.6} &\textbf{74.5} &\textbf{131.4} & \\
\midrule[1.0pt]
\end{tabular}
}
\end{center}
\caption{Overall performance of MS COCO online testing.}% where B-1, B-4, M and C are short for BLEU-1, BLEU-4, METEOR and CIDEr-D, respectively. * means ensemble model.  Numbers in \textbf{bold} denote the best performance in each column.} %The difference of all metrics between \emph{NIC+WC+WA+RL} and those models whose name begin with \emph{NIC} except \emph{NIC+WC(GT)} is statistically significant($p<0.05$)}.
\label{leaderboard}
\end{table}

\subsection{Experiment Setup}
\paragraph{Offline and Online Evaluation.} We evaluate our proposed model on MS COCO~\cite{lin2014microsoft}. Each image contains 5 human annotated captions. We split the dataset following~\cite{karpathy2015deep} with 113,287 images in the training set and 5,000 images in validation and test sets respectively. Besides, we test our model on MS COCO online testing datasets (40,775 images). The online testing has two settings, namely c5 and c40, with different numbers of reference sentences for each image ( 5 in c5 and 40 in c40).
\paragraph{Single and Ensemble Models.} Following the common practice of model ensemble in~\cite{yao2018exploring,yang2019auto,li2019entangled}, we build the ensemble version of \emph{TCIC} through averaging the output probability distributions of multiple independently trained instances of models. We use ensembles of two instances, and they are trained with different random seeds.
\paragraph{Evaluation Metrics.} We use BLEU~\cite{papineni2002bleu}, METEOR~\cite{banerjee2005meteor}, ROUGE-L~\cite{lin2003automatic}, CIDEr~\cite{vedantam2015cider}, and SPICE~\cite{anderson2016spice} as evaluation metrics, which are provided in COCO Caption Evaluation\footnote{\url{https://github.com/tylin/coco-caption}}.

\paragraph{Models in Comparison.} We compare our model with some state-of-the-art approaches. % \emph{NIC}~\cite{vinyals2015show} is the baseline CNN-RNN model trained with cross-entropy loss. \emph{SGAE}~\cite{yang2019auto} employs a pretrained sentence scene graph auto-encoder to model language prior, which better guide the caption generation from image scene graph. \emph{AoA}~\cite{huang2019attention}, \emph{MMT}~\cite{cornia2020m2}
\begin{enumerate}[-]
\itemsep-0.15em
    \item \emph{NIC}~\cite{vinyals2015show} is the baseline CNN-RNN model trained with cross-entropy loss.
    \item \emph{Up-down}~\cite{anderson2018bottom} uses a visual attention mechanism with two-layer LSTM, namely, top-down attention LSTM and language LSTM.
    \item \emph{GCN-LSTM}~\cite{yao2018exploring} presents Graph Convolutional Networks (GCN) to integrate both semantic and spatial object relations for better image encoding.
    \item \emph{SGAE}~\cite{yang2019auto} employs a pretrained sentence scene graph auto-encoder to model language prior, which better guide the caption generation from image scene graph.
    \item \emph{TOiW}~\cite{NIPS2019_9293} incorporates the object spatial relations to self-attention in Transformer.
    \item \emph{HIP}~\cite{yao2019hierarchy} models a hierarchy from instance level (segmentation), region level (detection) to the whole image.
    \item \emph{MMT}~\cite{cornia2020m2} learns a prior knowledge in each encoder layer as key-value pairs, and uses a mesh-like connectivity at decoding stage to exploit features in different encoder layers.
    % \item \emph{ETA}~\cite{li2019entangled} utilizes transformer to exploit text concepts and visual features simultaneously.
    \item \emph{TCIC} is our proposed model.
\end{enumerate}

\subsection{Implementation Details}
\noindent We utilize Faster-RCNN~\cite{ren2015faster} as the object detector and build the relation classifier following~\citet{zellers2018neural}. On top of these two components, we build a scene graph\footnote{\url{https://github.com/yangxuntu/SGsr}} for each image as the input of \emph{TTN-V}. We prune the vocabulary by dropping words appearing less than five times. Our encoder has 3 layers and the decoder has 1 layer, the hidden dimension is 1024, the head of attention is 8 and the inner dimension of feed-forward network is 2,048. The number of parameters in our model is 23.2M. The dropout rate here is 0.3. We first train our proposed model with cross-entropy with 0.2 label smoothing, $(\lambda_{1},\lambda_{2})=(0.5, 10.0)$ for 10k update steps, 1k warm-up steps, and then train it with reinforcement learning for 40 epochs, 40k update steps, $K$ in Eq.~(\ref{rl}) is 5. We use a linear-decay learning rate scheduler with 4k warm-up steps, the learning rates for cross-entropy and reinforcement learning are 1e-3 and 8e-5, respectively. The optimizer of our model is Adam~\cite{kingma2014adam} with (0.9, 0.999). The maximal region numbers per batch are 32,768 and 4,096. During decoding, the size of beam search is 3 and the length penalty is 0.1. 

\begin{figure}%[htbp]
\centering
\includegraphics[width=0.43\textwidth]{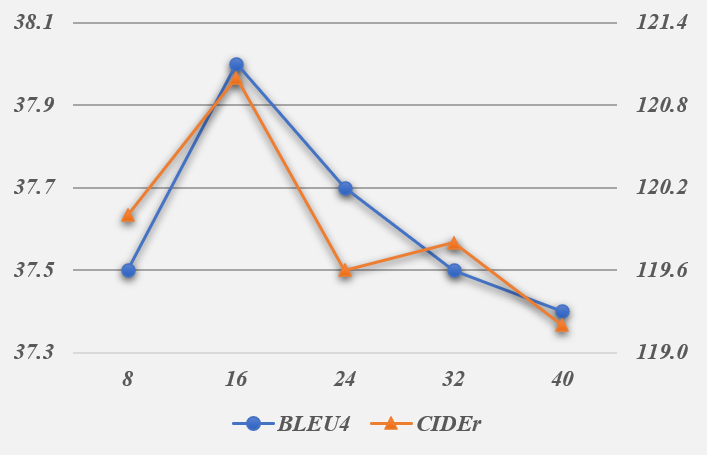}
\centering
\caption{The performance of \emph{TCIC} with different numbers of theme nodes (from 8 to 40, x-axis) in terms of CIDEr (right y-axis) and BLEU-4 (left y-axis).}
\label{vsn_scores}
\end{figure}

\subsection{Overall Performance}

\begin{table}
\begin{center}
%\textcolor{blue}{}%
\resizebox{0.46\textwidth}{!}{
\begin{tabular}{l|cccccc}
\midrule[1.0pt]
 &B-1 &B-4 &M &R &C &S \\
\midrule[1.0pt]
\emph{T+$\mathcal{O}$} &75.5 &34.6 &27.8 &56.0 &113.2 &21.0\\
%\emph{T+($\mathcal{O}$,$\mathcal{R}$)} 
+$\mathcal{R}$ &76.3 &35.6 &27.9 &56.4 &115.2 &21.0 \\
%\emph{T+($\mathcal{O}$,$\mathcal{R}$,$\mathcal{V})$} 
+$\mathcal{V}$ &76.9 &36.2 &28.1 &56.7  &117.6 &21.1\\
% \emph{T+($\mathcal{O}$,$\mathcal{R}$,$\mathcal{V})$+GE} 
+\emph{GE}&77.1 &36.9 &28.2 &56.7 &118.8 & 21.3 \\
%\emph{T+($\mathcal{O}$,$\mathcal{R}$,$\mathcal{V}$)+GE+SRG} 
+\emph{CR}&77.4 &37.4 &28.4 &57.3 &119.1 &21.5 \\
% \emph{TTN} 
+\emph{TA}&\textbf{78.1} &\textbf{38.3} &\textbf{28.4} &\textbf{58.0} &\textbf{121.0} &\textbf{21.6} \\
\midrule[1.0pt]
\end{tabular}
}
\end{center}
\caption{Ablation study for \emph{TCIC}. Components are added on top of the previous setting one by one from the first row to the bottom one.} %\emph{T}, \emph{GE}, \emph{SR} and \emph{TA} are Transformer, Group Embedding, caption re-construction and Theme Nodes Alignment.} %B-4, M, R, C and S are short for BLEU-4, METEOR, ROUGE, CIDEr-D and SPICE, respectively. Numbers in \textbf{bold} denote the best performance in each column.}
\label{ablation_study}
\end{table}

\noindent We present the performance of offline testing in Table~\ref{OverallPerformance} with two configurations of results (Cross Entropy and RL). One is trained with cross-entropy loss and the other is further trained via reinforce algorithm using CIDEr score as the reward. For single models, \emph{TCIC} achieves the highest scores among all compared methods in terms of most metrics (except BLEU-1 SPICE in RL version). For ensemble models, \emph{TCIC$^{\Sigma}$} outperforms other models in all metrics (except SPICE in RL version). We also evaluate our ensemble model on the online MS COCO test server and results are shown in Table~\ref{leaderboard}. \emph{TCIC$^{\Sigma}$} also generates better results compared to other three models. This validates the robustness of our model.

% \begin{table}
% \begin{center}
% %\textcolor{blue}{}%
% \small
% \begin{tabular}{l|ccccc}
% \midrule[1.0pt]
%  &B-4 &M &R &C &S \\
% \midrule[1.0pt]
% \emph{TTN}(-$\mathcal{R}$)(+$\mathcal{S}$) &37.3 &28.2 &57.3 &119.0 &21.3 \\
% \emph{TTN}(-$\mathcal{R}$)(+5$\mathcal{S}$) &\textbf{37.6} &\textbf{28.5} &\textbf{57.5} &\textbf{120.1} &\textbf{21.6} \\
% \midrule[1.0pt]
% \emph{TTN}(+$\mathcal{S}$) &37.4 &28.3 &57.3  &119.7 &21.4 \\
% \emph{TTN}(+5$\mathcal{S}$) &\textbf{38.3} &\textbf{28.4} &\textbf{58.0}  &\textbf{121.0} &\textbf{21.6} \\
% \midrule[1.0pt]
% \end{tabular}
% \end{center}
% \caption{Results of \emph{TTN} and\emph{TTN(-$\mathcal{R}$)} with $\mathcal{S}$ and $5\mathcal{S}$. B-4, M, R, C and S are short for BLEU-4, METEOR, ROUGE, CIDEr-D and SPICE, respectively. Numbers in \textbf{bold} denote the best performance in each column.}
% \label{influence_s}
% \end{table}
\label{Section_Case_Study}
\begin{figure}%[htbp]
\centering
\includegraphics[width=0.48\textwidth]{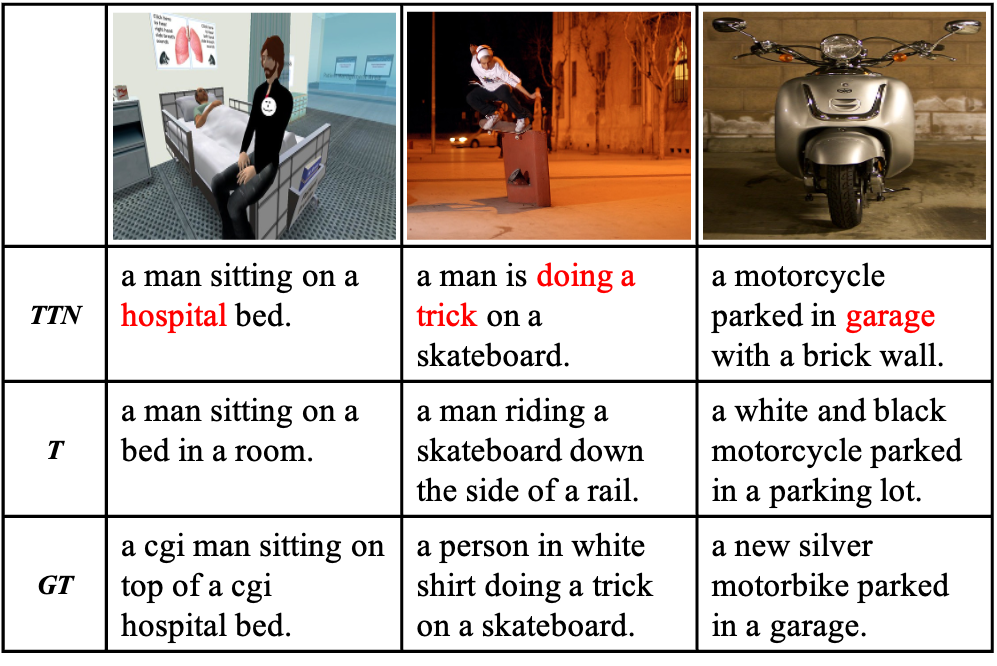}
\centering
\caption{Sample captions from \emph{TCIC} and Transformer.}
\label{case_study}
\end{figure}

\subsection{Ablation Study}
\noindent We perform ablation studies of our model. $\mathcal{O},\mathcal{R}, \mathcal{V}$ stand for nodes of object, relation and theme concept vectors, respectively. \emph{T}, \emph{GE}, \emph{CR} and \emph{TA} are used to denote transformer, group embedding, caption re-construction and theme nodes alignment, respectively. 

We add components into the basic setting one by one to track the effectiveness of the added component. We list results in Table~\ref{ablation_study}. We can see that the performance increases as components are added one by one. This demonstrates the effectiveness of different components. %Several findings stand out:
%\begin{enumerate}[-]
%\itemsep-0.15em
%    \item We can see that the performance increases as components are added one by one. This demonstrates the effectiveness of different components.
%    \item \emph{T}+$\mathcal{O}$+$\mathcal{R}$+$\mathcal{V}$ generates better results than  \emph{T}+$\mathcal{O}$+$\mathcal{R}$, which shows that theme nodes are able to capture more concept knowledge on top of scene graph.
%    \item +\emph{CR}+\emph{TA} improves 2.2 points in terms of CIDEr and 1.4 points in terms of BLEU-4. This indicates the effectiveness of using caption re-construction task for theme concepts learning. And \emph{TA} is effective in cross-modality semantic alignment.
%\end{enumerate}

%From Table~\ref{ablation_study}, models with different modules all outperforms the baseline Transformer. 

\section{Further Analysis}
% In this section, we further analyze the performance of modules involving theme concepts in depth. \
\noindent We qualify the influence of number of theme nodes in model performance in \S~\ref{Influence_of_Theme_Node_Number}, interpret the semantics of theme concepts in \S~\ref{Interpretation_of_Theme_Nodes}, and present case studies in \S~\ref{Case_Study}.

\subsection{Influence of Theme Node Number}
\label{Influence_of_Theme_Node_Number}
\noindent We investigate the influence of the number of theme nodes on the model performance in terms of CIDEr and BLEU-4. The results are shown in Figure~\ref{vsn_scores}. With the number of theme nodes increasing, both scores rise in the beginning, peak at 16 and go down after. This phenomenon indicates when the number of theme nodes is small, their modeling capacity is not strong enough to model theme concepts in the image for caption generation. When the number of theme nodes gets larger, different theme nodes may conflict with each other, which hurts the performance.

\subsection{Interpretation of Theme Nodes}
\label{Interpretation_of_Theme_Nodes}
% \subsection{Concretization of Theme Nodes} % 这个题目会不会更好一点？
% Theme Node is proposed to dig the pattern of theme in scene graph, thus we check those elements in encoder and decoder that attend to these theme nodes to display the characteristics of different theme nodes. In encoder-layer\#3, for each object node $o_{j}$, we are able to obtain its attention weights to these Theme Nodes through Equation~\ref{new-self-attention} and Equation~\ref{group-self-attention}, if some $v_{k}$ are the top2 keys, then $\#\{v_{k}\gets o_{j}\}+1$. The larger $\#\{v_{k}\gets o_{j}\}$ of some $o_{j}$ is, we consider that $o_{j}$ is more related to $v_{k}$. At last, for visualization, we pick up 3 theme nodes and select 10 out of the 20 most frequently appeared objects based on $\#\{v_{k}\gets o_{j}\}$. So as for the encoder-decoder-attention in decoder-layer\#1. The result is shown in Table~\ref{attention_visualization}.
\noindent Theme concepts are introduce to represent cross modality high-level semantics. Through linking theme nodes with objects in \emph{TTN-V} and words in \emph{TTN-L} based on attention scores, we try to visulize semantics of these theme nodes.

In \emph{TTN-V}, for each object node $o_{i}$, we obtain its attention scores to theme nodes through $\textbf{SAN}$. If the attention score to $v_{j}$ ranks in the top-2, we link $v_{j}$ with $o_{i}$ and increase their closeness scores by 1. We pick about 8 objects out of top-20 for each theme node in terms of closeness score. Same procedures are done in \emph{TTN-L} to link theme nodes and words through $\textbf{SAN}$. 
We list 3 theme nodes and their relevant objects and words in Table~\ref{attention_visualization}. 
We conclude that:
\begin{enumerate}[-]
\itemsep-0.2em
\item Different theme nodes are related to different categories of objects and words. node\#1 is clothing, while node\#2 is related to transportation. This indicates different theme nodes contain different high-level semantics.

% \item In decoder, theme nodes not only connect to nouns, but also verbs under similar theme, such as ``transit" in node\#2 and ``fries" in node\#3. % \item Objects have different preference on theme nodes, and these are some clustering phenomenon of these preference. As we can see, node\#1 may be more favored by clothing, but node#2 is more related to transportation. This indicates different nodes cover different aspects of high-level theme concepts.
\item There exists correlation between \emph{TTN-V} and \emph{TTN-L}. Theme node\#3 in encoder and decoder are both related to food. This reveals that \emph{TTN} is able to align semantics of vision and language to some extent. 
% \item The same node would be favored by different categories of objects across different layers. For exƒample, node\#2 is more related to indoors items in Layer\#1, and also some animals in Layer#3. This is because our theme node would dynamically adapt its representation according to different source elements.  

% \item One node is able to connect with different object categories. For example, node\#4 in Layer\#3 is attend to different crowds and clothes. This demonstrates our model is capable of modeling relationship between different object categories in a latent manner.
\end{enumerate}

\begin{table}[tb]
\begin{center}
\resizebox{0.48\textwidth}{!}{
\begin{tabular}{c|c|c}
\midrule[1.0pt]
&\emph{TTN-V} &\emph{TTN-L} \\
% \multirow{2}{*}{Scene Graph} &\multicolumn{3}{c}{Vocabulary Size} \\
% &#Object &#Attribute &#Relation \\
% \midrule[1.0pt]  
\midrule[1.0pt]
\multirow{3}{*}{1}&jacket, tie, jean, &shorts, shirt, attire, \\
&sneaker, belt, vest, &mitt, hats, clothes, \\
&short, uniform, hat &uniforms, clothing, suits \\
\midrule[0.5pt]
\multirow{3}{*}{2}&wheel, pavement, car, &vehicles, helmets,\\
&van, sidewalk, truck, &carriages, passengers,   \\
&street, road, bus & transit, tracks, intersection  \\
\midrule[0.5pt]
\multirow{3}{*}{3}&vegetable, tomato, fruit, &fries, slices, chips, \\
&onion, orange, broccoli, &eat, food, vegetables, \\
&banana, apple, meat &plates, pizzas, sandwiches \\
\midrule[1.0pt]
\end{tabular}
}
\end{center}
\caption{The most related elements in TTN-V encoder and TTN-L encoder for theme node\#1, \#2 and \#3.}
\label{attention_visualization}
\end{table}

\subsection{Case Study}
\label{Case_Study}
\noindent We show sample captions generated by \emph{T} (transformer) and \emph{TCIC} in Figure~\ref{case_study}. \emph{T} only describes the low-level facts of images, but \emph{TCIC} infers high-level concepts (words in red) on top of these facts such as, ``hospital", ``doing a trick" and ``garage".

\section{Conclusion and Future Work}
\label{SectionConclusion}
\noindent In this paper, we explore to use theme concepts to represent high-level semantics cross language and vision. Theme concepts are modeled as memory vectors and updated inside a novel transform structure TTN. We use two tasks to enable the learning of the theme concepts from both images and captions. On the vision side, TTN takes both scene graph based features and theme concept vectors for image captioning. On the language side, TTN takes textual features and theme concept vectors for caption re-construction. Experiment results show the effectiveness of our model and further analysis reveals that TTN is able to link high-level semantics between images and captions. 
In future, we would like to explore the interpretation of theme concept vectors in an explicit way. Besides, the application of theme concept vectors for other downstream tasks would also be an interesting direction to explore. 

\section*{Acknowledgements}
\noindent This work is partially supported by Ministry of Science and Technology of China (No.2020AAA0106701), Science and Technology Commission of Shanghai Municipality Grant (No.20dz1200600, 21QA1400600).

\newpage

\bibliographystyle{named}
\bibliography{ijcai21}

\end{document}